\documentclass[conference,compsoc]{IEEEtran}

\usepackage{graphicx}
\usepackage{amsmath}
\usepackage{multirow} 
\usepackage{tikz}
\usepackage{fontawesome5}
\usepackage{pifont}
\usepackage{enumitem}
\usetikzlibrary{shapes.geometric, arrows.meta, positioning, fit, calc, decorations.pathreplacing}
\usetikzlibrary{
    shapes.geometric, 
    arrows.meta,      %
    positioning,      %
    fit,              %
    calc,             %
    decorations.pathreplacing %
}
\usetikzlibrary{positioning}
\usepackage{url}
\usepackage{hyperref}
\renewcommand{\paragraph}[1]{\textbf{#1.}}
\usepackage{graphicx}   %
\usepackage{lipsum}     %
\usepackage{float}

\ifCLASSOPTIONcompsoc
  \usepackage[nocompress]{cite}
  \usepackage{amsmath}
  \usepackage{amssymb}
  \usepackage{booktabs}
\usepackage{algorithm}
\usepackage{algpseudocode}
\usepackage{etoolbox}
\usepackage{pgfplots}
\pgfplotsset{compat=1.18}
\usepgfplotslibrary{colormaps}

\makeatletter
\renewcommand{\Return}[1]{\State \algorithmicreturn\ #1}
\makeatother

\else
  \usepackage{cite}
\fi

\ifCLASSINFOpdf
\else
\fi

\begin{document}
\title{OmniOCR: Generalist OCR for Ethnic Minority Languages}

\author{
    \textbf{Bonan Liu}$^{1}$\quad
    \textbf{Zeyu Zhang}$^{2*}$\quad 
    \textbf{Bingbing Meng}$^{1}$\quad 
    \textbf{Han Wang}$^{1}$\quad
    \textbf{Hanshuo Zhang}$^{1}$\\
    \textbf{Chengping Wang}$^{1}$\quad
    \textbf{Daji Ergu}$^{1}$\quad
    \textbf{Ying Cai}$^{3\dag}$ \vspace{0.1cm}\\
    $^1$Southwest Minzu University\quad
    $^2$AI Geeks\vspace{0.05cm}\\
    \small $^*$Project lead. $^\dag$Corresponding author: caiying34@yeah.net.
}

\maketitle

\begin{abstract}
Optical character recognition (OCR) has advanced rapidly with deep learning and multimodal models, yet most methods focus on well-resourced scripts such as Latin and Chinese. Ethnic minority languages remain underexplored due to complex writing systems, scarce annotations, and diverse historical and modern forms, making generalization in low-resource or zero-shot settings challenging. To address these challenges, we present OmniOCR, a universal framework for ethnic minority scripts. OmniOCR introduces Dynamic Low-Rank Adaptation (Dynamic LoRA) to allocate model capacity across layers and scripts, enabling effective adaptation while preserving knowledge.A sparsity regularization prunes redundant updates, ensuring compact and efficient adaptation without extra inference cost. Evaluations on TibetanMNIST, Shui, ancient Yi, and Dongba show that OmniOCR outperforms zero-shot foundation models and standard post training, achieving state-of-the-art accuracy with superior parameter efficiency, and compared with the state-of-the-art baseline models, it improves accuracy by 39\%–66\% on these four datasets.
Code: \url{https://github.com/AIGeeksGroup/OmniOCR}.
\end{abstract}

\IEEEpeerreviewmaketitle

\begin{figure}
\centering
\begin{tikzpicture}
\begin{axis}[
    ybar,
    width=0.48\textwidth,
    height=8cm,
    ylabel={Accuracy (\%)},
    ymin=0, ymax=100,
    bar width=8pt,
    xtick=data,
    xticklabels={
        GPT-4o,
        Qwen-VL-Max,
        GLM-4v-Plus,
        Gemini 2.5 Pro,
        DeepSeek-VL2,
        InternVL3-78B,
        Moonshot-v1,
        Kimi-VL,
        Pixtral Large,
        RolmOCR,
        Doubao-1.5-Vision-Pro,
        Claude-3.7-Sonnet,
        Qwen-VL-OCR,
        RolmOCR (LoRA),
        RolmOCR (Full FT),
        OmniOCR (Ours)
    },
    xticklabel style={font=\scriptsize, rotate=45, anchor=east},
    tick label style={font=\scriptsize},
    ylabel style={font=\scriptsize},
    nodes near coords,
    every node near coord/.append style={font=\scriptsize, rotate=90, anchor=west,color=black},
    enlarge x limits=0.08,
    axis line style={line width=1pt},
    tick style={line width=1pt},
    axis x line*=bottom,
    axis y line*=left,
    point meta=y
]
\addplot+[ybar, draw=black, fill] 
coordinates {
    (0,25.61) (1,26.15) (2,27.24) (3,27.41) (4,27.83) (5,30.82) 
    (6,28.84) (7,29.45) (8,30.56) (9,29.31) (10,32.55) (11,34.63) (12,38.85) 
    (13,80.52) (14,89.21) (15,90.37)
};
\end{axis}
\end{tikzpicture}
\caption{Accuracy comparison on Tibetan dataset across all models.}
\label{fig:tibetan_acc_bar_orange}
\end{figure}
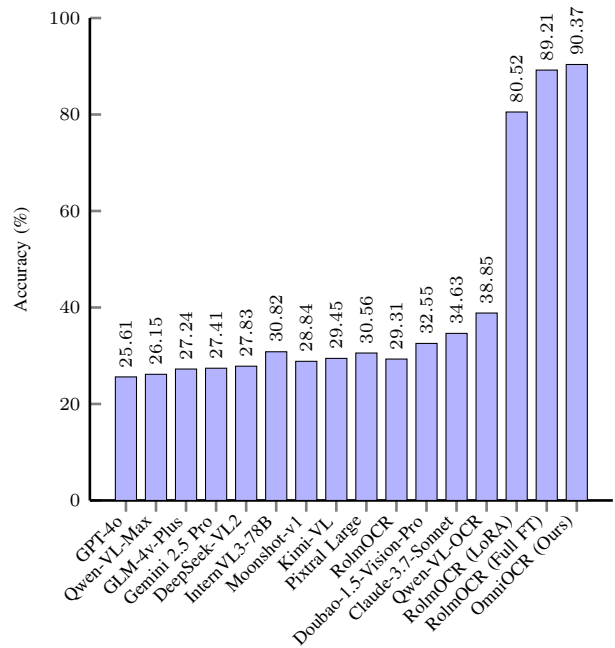

\section{Introduction}

OCR has achieved remarkable progress with deep learning and, more recently, large multimodal models. However, most existing methods target well-resourced scripts such as English or Chinese, while ethnic minority languages remain underexplored. Their complex writing systems, limited annotations, and coexistence of diverse historical and modern forms,pose unique challenges to conventional OCR.

Early OCR systems for ethnic minority scripts relied on handcrafted features and script-specific segmentation strategies, such as those designed for Mongolian and Tibetan \cite{peng2006multilingual, drup2010study}. With segmentation becoming a bottleneck, research shifted to segmentation-free deep learning, including CNN-based sliding window models and attention-based sequence models \cite{zhang2021ocr, zheng2018segmentation, zhang2017segmentation}. Recently, vision-language models (VLMs) and large language models (LLMs) have demonstrated strong cross-modal representation capabilities \cite{chung2025finetuning, li2023trocr, greif2025multimodal}, yet still struggle to generalize to ethnic minority scripts, particularly in low-resource or zero-shot settings.

To address these challenges, we propose \textbf{OmniOCR}, a universal framework for ethnic minority language OCR. Built on RolmOCR \cite{bai2025qwen2}, OmniOCR integrates \textbf{Dynamic LoRA}, which adaptively allocates model capacity across layers and scripts. This design enables effective adaptation to structurally complex scripts while mitigating overfitting in low-resource scenarios. A sparsity regularization further prunes redundant updates, ensuring compact adaptation without extra inference cost.
We evaluate OmniOCR on four representative datasets—TibetanMNIST, Shui, ancient Yi, and Dongba. Experiments show consistent gains over zero-shot foundation models and standard post training, achieving state-of-the-art performance with superior parameter efficiency.Taking the Tibetan dataset as a key verification scenario,the accuracy results are shown in \textbf{Figure \ref{fig:tibetan_acc_bar_orange}}. 

In summarization, our contribution of our paper can be summarized in 3 folds:
\begin{itemize}[noitemsep, topsep=0pt]  
    \item We introduce \textbf{OmniOCR}, the first universal OCR framework for heterogeneous ethnic minority scripts.
    \item We design a \textbf{Dynamic LoRA} module that balances knowledge retention and efficient adaptation across scripts.
    \item We establish new benchmarks on four ethnic minority language datasets, surpassing existing baselines in both accuracy and efficiency, and improving accuracy by 39\%–66\% on these four datasets.
\end{itemize}

\section{Related Work}

\paragraph{LLMs in OCR}
In recent years, OCR research has shifted from task-specific models to general-purpose large models. LLMs, especially multimodal ones (MLLMs), show strong contextual and visual capabilities, offering new solutions to long-standing OCR challenges. Even before LMMs, the community had moved from CNN-RNNs to Transformer-based frameworks. For instance, Li et al.\cite{li2023trocr} proposed TrOCR, combining pretrained image and text Transformers for end-to-end recognition, achieving strong results on printed, handwritten, and scene text.The rise of MLLMs further enhanced OCR performance. Greif et al.\cite{greif2025multimodal} showed that Gemini 2.0 Flash outperformed traditional OCR systems without fine-tuning. Benchmarks such as OCRBench\cite{liu2024ocrbench}, CC-OCR\cite{yang2024cc}, and Reasoning-OCR\cite{he2025reasoning} systematically evaluated MLLMs across document parsing, key information extraction, multilingual recognition, and reasoning. However, results vary across languages: Sohail et al.\cite{sohail2024deciphering} found GPT-4o struggled on complex scripts like Urdu and Tajik, performing worse as text length increased.To overcome these issues, specialized OCR-oriented LMMs have emerged. Chen et al.\cite{chen2025ocean} introduced Ocean-OCR, using NaViT to handle variable-resolution inputs and trained on large OCR datasets, achieving strong results across scene, document, and handwriting recognition. 

\paragraph{OCR for  Ethnic Minority Languages}
In the early development of OCR for ethnic minority languages, most methods relied on accurate character segmentation. Peng et al.\cite{peng2006multilingual} built a printed Mongolian system requiring segmentation for its vertical layout, while Drup et al.\cite{drup2010study} improved Tibetan OCR with adaptive binarization and connected-component analysis. Sun et al.\cite{sun2019yi} applied Tesseract to Yi script but still needed repeated box generation with jTessBox Editor. As segmentation became a bottleneck, later studies shifted to segmentation-free deep learning. Zhang et al.\cite{zhang2021ocr} developed a CNN-based sliding-window framework for Manchu; Zheng et al.\cite{zheng2018segmentation} improved printed Manchu OCR with a nine-layer CNN; and Zhang et al.\cite{zhang2017segmentation} introduced an attention-based seq2seq model for Mongolian. More recently, vision-language models (VLMs) have emerged. Chung et al.\cite{chung2025finetuning} fine-tuned open-source VLMs (e.g., LLaMA, Qwen) on 60,000 synthetic Manchu word images, achieving 93.1\% accuracy on real manuscripts, surpassing CRNN baselines

\begin{table}[t]
\centering
\caption{STATISTICS OF THE FOUR ETHNIC MINORITY LANGUAGE DATASETS USED FOR TRAINING AND EVALUATION OF OMNIOCR.}
\label{tab:dataset_stats} 
\resizebox{\linewidth}{!}{
\begin{tabular}{l|c|c|c|c}
\toprule
\textbf{Dataset} & \textbf{Classes} & \textbf{Samples} & \textbf{Script Type} & \textbf{Image Size} \\
\midrule
TibetanMNIST \cite{yuan2018tibetanmnist} & 10 & 17,768 & Digits & 111M \\
\hline
Shui Dataset \cite{Yang2023ShuiOCR} & 12 & 5,280 &Pictographic & 21.6M \\
\hline
Ancient Yi Script \cite{liu2024ancient} & 30 & 10,840 & Logographic & 62.2M \\
\hline
Dongba Script \cite{luo2023multiple} & 30 & 14,906 & Pictographic & 20.2M \\
\bottomrule
\end{tabular}}
\end{table}
\section{Datasets}
To train and evaluate the OmniOCR model, we curated four publicly available datasets encompassing diverse ethnic minority scripts, covering both historical and contemporary writing systems as well as handwritten numerals. Detailed statistics of these datasets—including the number of classes, total samples, script type, and image size—are summarized in \textbf{Table \ref{tab:dataset_stats}}.

\textbf{TibetanMNIST}\cite{yuan2018tibetanmnist} is a benchmark dataset for handwritten Tibetan digits, which plays a crucial role in advancing OCR research for the Tibetan script by filling a gap in publicly available resources. It contains 17,768 images of handwritten Tibetan numerals, produced by multiple researchers from Tibetan studies institutes, ensuring a wide variety of writing styles. By incorporating this dataset, we can rigorously assess the proposed OmniOCR model’s ability to generalize across variations in character morphology and its effectiveness in recognizing complex, non-alphabetic scripts.

\textbf{Shui Dataset}\cite{Yang2023ShuiOCR} is a dataset of ancient Shui characters, comprising 5,280 images across 12 representative classes. These images depict elements of the natural and cultural world, including mountains, rivers, trees, the sun, the moon, stars, animals, deities, and divination symbols, reflecting the richness of Shui cultural heritage. This dataset enables assessment of OmniOCR’s ability to recognize complex pictographic scripts.

\textbf{Ancient Yi Script Handwritten Character Dataset}\cite{liu2024ancient} is a dataset of handwritten ancient Yi script, originally comprising 2,922 character classes with over 427,000 samples. To manage computational complexity and focus the recognition experiment, we selected a representative subset of 30 classes based on two criteria: high frequency in Yi documents, ensuring sufficient samples per class, and diversity in stroke structures, capturing variations in handwriting styles. For each selected class, an equal number of images were randomly sampled from multiple writers to mitigate class imbalance and maintain evaluation reliability. This enables a rigorous assessment of OmniOCR’s generalization capability on handwritten Yi characters.

\textbf{The Handwritten Dongba Character Dataset}\cite{luo2023multiple} is a large-scale collection of single Dongba characters, constructed through manual imitation, character cropping, grayscale conversion, binarization, and size normalization. The original dataset comprises 1,404 Dongba character classes, totaling 445,273 images and covering 2,546 variant forms, with each class containing 103 to 1,091 samples. To ensure both experimental feasibility and representativeness, we filtered the dataset based on the characters’ pictographic recognizability (i.e., how easily the character shape can be identified) and practical usage frequency. As a result, we selected the 30 highest-quality character classes, forming a subset that balances recognizability and representativeness, which is used to rigorously evaluate OmniOCR’s performance on Dongba character recognition.

\begin{figure*}[t]
    \centering
    \includegraphics[width=\textwidth]{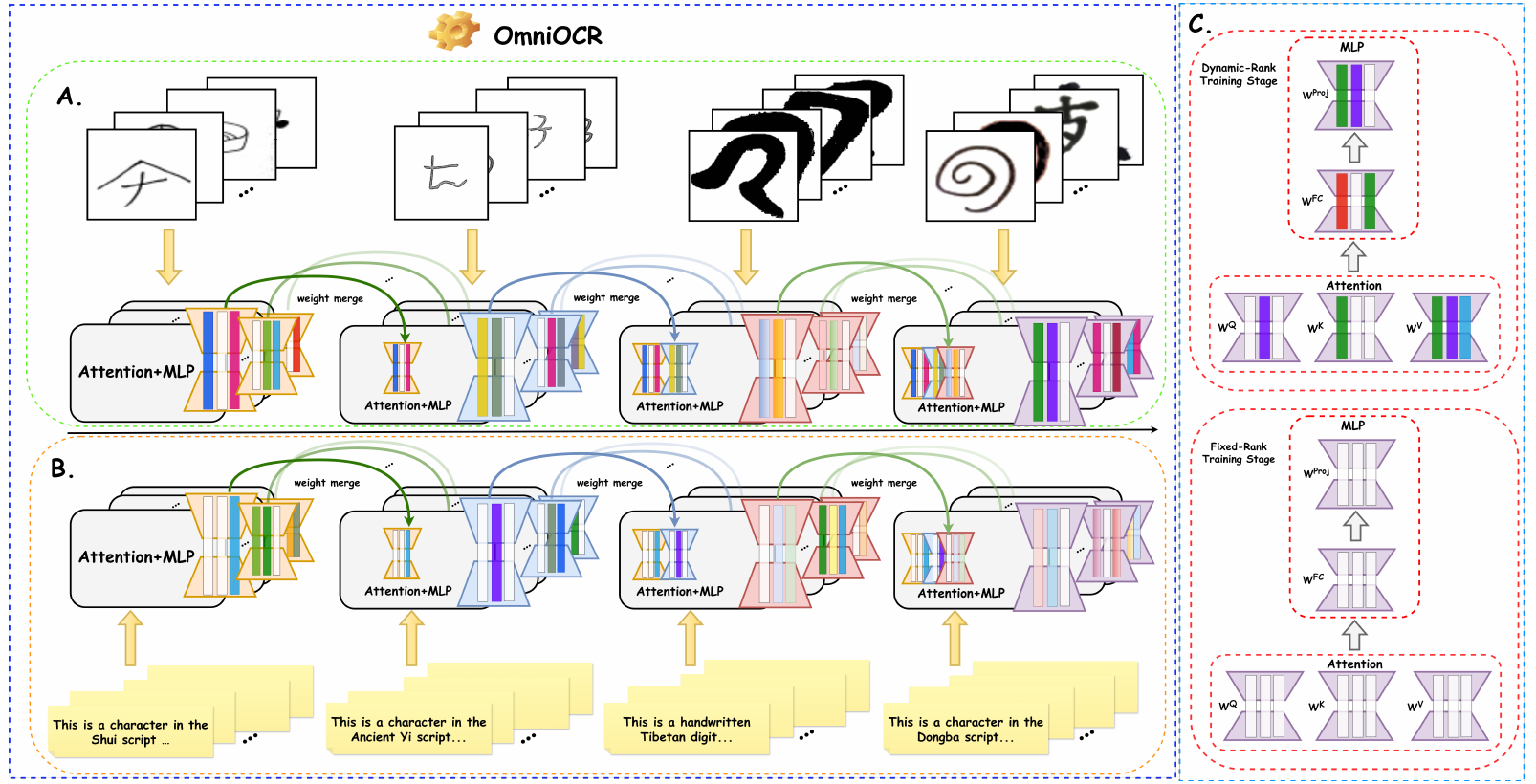}
    \caption{\textbf{OmniOCR.} A represents the processing procedure of the model's Vision Encoder; B represents the processing procedure of the model's Text Encoder; C respresents two distinct parameter-efficient fine-tuning methods: Dynamic-Rank Training and Fixed-Rank Training.}
    \label{fig:omniocr_architecture}
\end{figure*}

\section{Method}

\subsection{Overview}
In this work, we propose OmniOCR, a generalist OCR framework tailored for ethnic minority languages. Our approach builds upon the vision-language foundation model RolmOCR\cite{bai2025qwen2}, which provides strong cross-lingual representation capability. However, directly applying RolmOCR to low-resource minority scripts often suffers from limited recognition accuracy due to script-specific structural variations and severe data scarcity. 
Inspired by \cite{lu2024adaptive}, we design a dynamic LoRA module within OmniOCR that adaptively adjusts the rank across different layers, thereby balancing the acquisition of new knowledge with the retention of previously learned scripts.The complete architecture of OmniOCR—including its Vision Encoder, Text Encoder, and the two parameter-efficient post training methods that underpin its performance—is illustrated in \textbf{Figure \ref{fig:omniocr_architecture}}.In addition, to further reduce GPU memory consumption during training, we preprocess the datasets by resizing and normalizing the input images while preserving script readability. We evaluate OmniOCR on four representative minority language datasets with diverse writing systems, and the results demonstrate that our framework achieves robust performance across underrepresented scripts while effectively mitigating catastrophic forgetting.

\begin{figure*}[t]
    \centering
    \includegraphics[width=\textwidth]{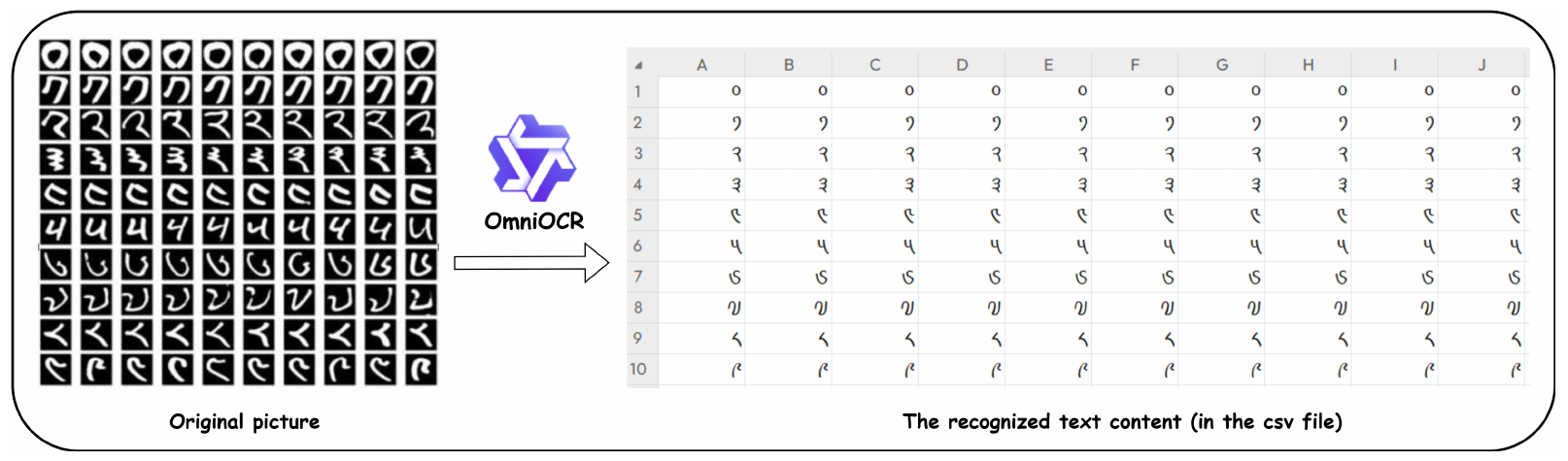}
    \caption{Demonstration of Recognition Performance for Tibetan Handwritten Digits via OmniOCR. The results show the accuracy and visual recognition effect of OmniOCR on the Tibetan handwritten digit dataset.}
    \label{fig:tibetan_digit_recognition}
\end{figure*}

\subsection{Dynamic LoRA Adaptation} 
Recognizing minority languages poses unique challenges compared to mainstream scripts, due to their heterogeneous writing systems (e.g., Tibetan’s distinct numeral forms, Ancient Yi’s pictographic symbols, Shui script’s water-based ideograms, and Dongba’s logographic structures) and the limited availability of annotated data. To address these challenges, we design OmniOCR, a unified framework that adapts a general OCR backbone to multiple minority scripts in a parameter-efficient manner.
A key component of OmniOCR is a dynamic LoRA module that tailors the model’s capacity to each script’s characteristics. Instead of applying a fixed-rank update, which may underfit complex scripts or overfit scarce data, we allow the update rank to be adaptively determined for each layer and task. Specifically, for a pre-trained weight matrix $W_{0}^{t,m}$ at task $t$ and layer $m$, the update $\Delta W^{t,m}$ is expressed as:
$$
\Delta W^{t,m} = \sum_{i=1}^{r} w_i^{t,m} B_i^{t,m} A_i^{t,m}
$$
where $r$ is the maximum candidate rank, $B_i^{t,m}$ and $A_i^{t,m}$ are low-rank matrices, and $w_i^{t,m}$ is a learnable importance weight. This formulation enables the model to allocate more capacity to scripts with complex visual structures (e.g., Dongba or Ancient Yi), while using fewer ranks for simpler ones (e.g., Tibetan digits), thereby achieving a balance between adaptability and efficiency.
To further enhance robustness in low-resource scenarios, we impose an $\ell_1$ sparsity regularization on the importance weights:
$$
\mathcal{L}^t_{\min} := \mathcal{L}^t_{\sup} + \lambda \sum_{m=1}^{M} \|w^{t,m}\|_1
$$
where $\mathcal{L}^t_{\sup}$ is the supervised loss, $M$ is the number of updated matrices, and $\lambda$ controls sparsity. This design encourages the model to retain only the most critical update directions while pruning redundant ones, ensuring compact adaptation without extra inference cost.The complete training procedure is outlined in \textbf{Algorithm \ref{alg:omniocr}}.
Through this mechanism, OmniOCR is able to efficiently adapt to the structural diversity across Tibetan, Ancient Yi, Shui, and Dongba scripts, while simultaneously mitigating catastrophic forgetting when learning sequentially across different minority languages. Moreover, by pruning redundant update directions and retaining only the most critical ones, the framework achieves compact adaptation without introducing additional inference overhead, making it both effective in low-resource scenarios and practical for real-world applications.

\begin{algorithm}[!htbp]
\caption{OmniOCR with Dynamic LoRA Adaptation}
\label{alg:omniocr}
\begin{algorithmic}[1]
\Require Pre-trained backbone $W_0$; max rank $r$; sparsity weight $\lambda$; datasets $\mathcal{D}=\{D_1,\dots,D_T\}$
\Ensure Adapted model with compact LoRA modules
\For{each task $t$ with dataset $D_t$}
    \State Freeze $W_0$, initialize $\{A_i^{t,m}, B_i^{t,m}, w_i^{t,m}\}$
    \For{mini-batch $\mathcal{B} \subseteq D_t$}
        \State Compute update:
        \[
            \Delta W^{t,m} = \sum_{i=1}^r w_i^{t,m} B_i^{t,m} A_i^{t,m}
        \]
        \State Forward with $W_0 + \Delta W$, compute  loss $\mathcal{L}_{\text{sup}}$
        \State Add sparsity:
        \[
            \mathcal{L} \gets \mathcal{L}_{\text{sup}} + \lambda \sum_m \|w^{t,m}\|_1
        \]
        \State Backpropagate and update only $\{A,B,w\}$
    \EndFor
    \State Prune update directions with small $|w_i^{t,m}|$
\EndFor

\Return Updated backbone $W_0$ with compact LoRA modules
\end{algorithmic}
\end{algorithm}

\begin{figure*}[t]
    \centering
    \includegraphics[width=\textwidth]{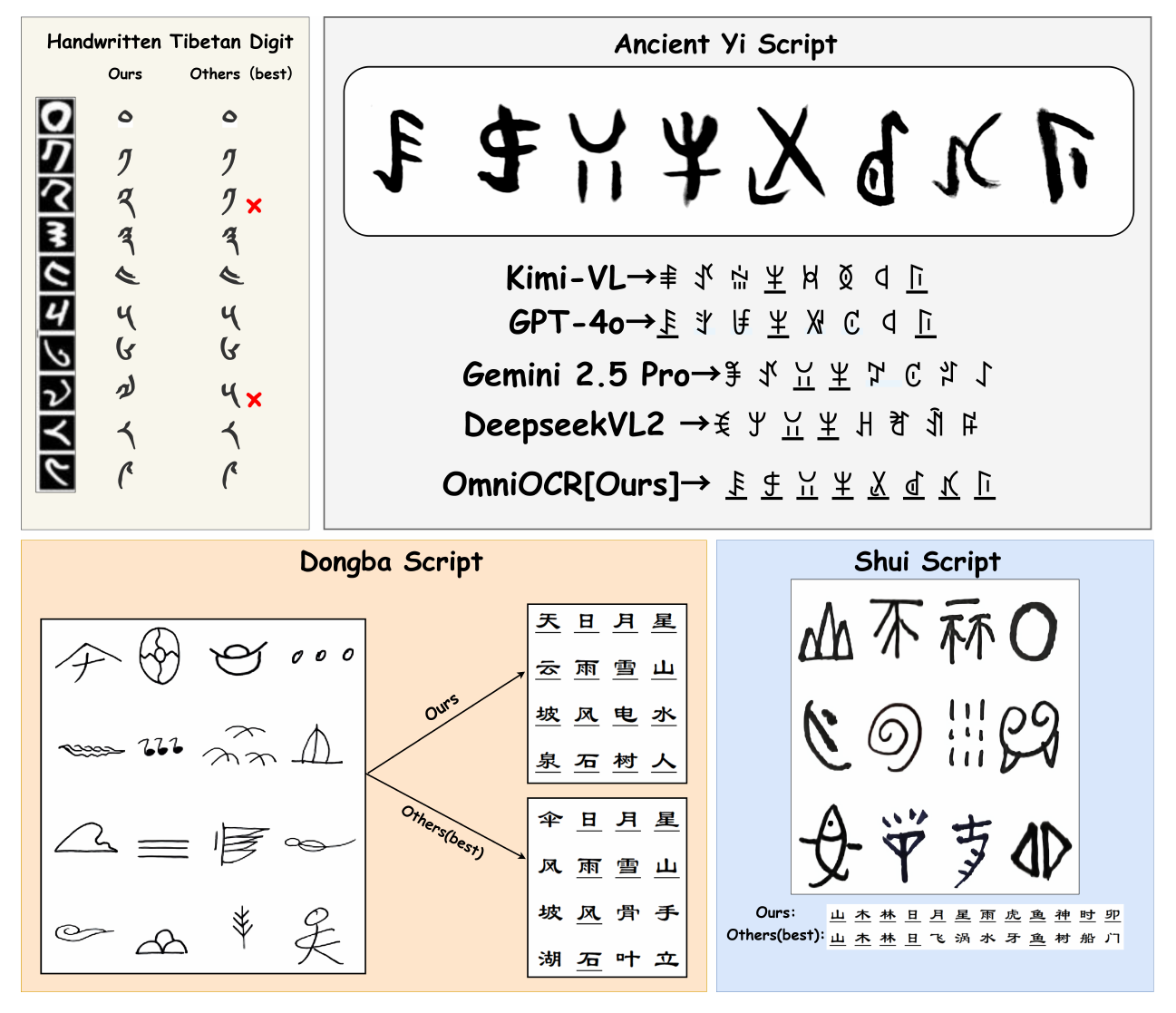}
    \caption{The performance of OmniOCR evaluated on four representative datasets—TibetanMNIST, Shui, Ancient Yi, and Dongba—highlighting its ability to generalize across heterogeneous scripts and writing systems.}
    \label{fig:omniocr_multi_dataset_perf}
\end{figure*}

\section{Experiments}

\subsection{Implementation Details}
We implement OmniOCR with PyTorch. 
All experiments are conducted on a server equipped with a single NVIDIA H20 GPU (96GB memory).. 
Input images are resized to $48\times48$ pixels. For training, standard data augmentations are applied, including random horizontal flip, random rotation ($\pm 10^\circ$), random resized crop (scale 0.8--1.0), and color jitter. 
Evaluation uses only resizing and normalization.

The base model is initialized from the pre-trained RolmOCR\cite{bai2025qwen2}. 
We replace selected linear layers (self-attention projections and MLP layers) with the proposed Dynamic LoRA modules, which support dynamic rank adaptation. 
The initial LoRA rank is set to $r=8$, LoRA scaling factor $\alpha=16$. 
During training, the effective rank is adjusted via soft-shrinkage on rank weights, encouraging sparsity while retaining task-relevant capacity. 
To prevent overfitting, learned low-rank updates are merged back into the frozen backbone at checkpoint saving.

We train with the AdamW optimizer, using an initial learning rate of $5\times10^{-6}$, weight decay $1\times10^{-2}$, and gradient clipping at 1.0. 
We adopt a batch size of $1$ (due to GPU memory constraints) and use gradient accumulation over 2 steps to achieve an effective batch size of $2$. 
Training runs for 30 epochs with early stopping based on validation accuracy. 
Mixed-precision training with BF16 is enabled to improve efficiency. 

The best model is selected based on validation accuracy and saved together with the processor for inference.

\begin{table*}[h]
\centering
\caption{PERFORMANCE (\%) ON FOUR ETHNIC MINORITY LANGUAGE DATASETS. ZERO-SHOT MODELS AND FINE-TUNING METHODS ON ROLMOCR ARE COMPARED.}
\resizebox{\textwidth}{!}{
\begin{tabular}{l|ccc|ccc|ccc|ccc}
\toprule
\multirow{2}{*}{Models / Methods} 
& \multicolumn{3}{c|}{Tibetan\cite{yuan2018tibetanmnist}} 
& \multicolumn{3}{c|}{Ancient Yi\cite{liu2024ancient}} 
& \multicolumn{3}{c|}{Shui Script\cite{Yang2023ShuiOCR}} 
& \multicolumn{3}{c}{Dongba Script\cite{luo2023multiple}} \\
\cmidrule(lr){2-4}\cmidrule(lr){5-7}\cmidrule(lr){8-10}\cmidrule(lr){11-13}
& Acc & Recall & F1 
& Acc & Recall & F1 
& Acc & Recall & F1 
& Acc & Recall & F1 \\
\midrule
\multicolumn{13}{c}{\textbf{Zero-Shot}} \\
\midrule
Kimi-VL\cite{team2025kimi}             & 29.45 & 28.73 & 28.96 & 19.32 & 20.14 & 19.92 & 49.68 & 48.91 & 49.15 & 32.33 & 33.07 & 32.83 \\
Moonshot-v1\cite{moonshot2025v1}       & 28.84 & 31.12 & 29.95 & 17.14 & 18.75 & 17.92 & 47.27 & 49.35 & 48.29 & 30.31 & 32.01 & 31.14 \\
Pixtral Large\cite{mistral2025pixtral}       & 30.56 & 25.89 & 27.98 & 20.13 & 25.78 & 22.73 & 51.42 & 48.35 & 49.81 & 34.27 & 30.15 & 32.08 \\
Doubao-1.5-Vision-Pro\cite{bytedance2025doubao}       & 32.55 & 31.42 & 31.76 & 22.34 & 21.23 & 21.51 & 54.49 & 55.17 & 54.88 & 30.72 & 29.88 & 30.14 \\
GLM-4v-Plus\cite{zhipu2025glm4vplus}          & 27.24 & 26.43 & 26.71 & 19.83 & 18.92 & 19.16 & 46.38 & 47.21 & 46.95 & 33.25 & 32.64 & 32.87 \\
DeepSeek-VL2\cite{wu2024deepseek}        & 27.83 & 32.95 & 30.21 & 22.58 & 18.32 & 20.21 & 52.76 & 57.94 & 55.26 & 35.81 & 32.47 & 34.08 \\
Qwen-VL-Max\cite{qwen2025vlmax}         & 26.15 & 27.32 & 27.06 & 21.23 & 20.47 & 20.73 & 36.22 & 35.69 & 35.89 & 33.11 & 34.28 & 34.04 \\
Qwen-VL-OCR\cite{qwen2025vlocr}         & 38.85 & 37.96 & 38.18 & 14.53 & 15.31 & 15.10 & 16.03 & 15.28 & 15.49 & 11.15 & 11.84 & 11.62 \\
InternVL3-78B\cite{zhu2025internvl3}       & 30.82 & 32.67 & 31.71 & 19.26 & 21.04 & 20.12 & 53.91 & 55.62 & 54.75 & 31.43 & 33.14 & 32.26 \\
Gemini 2.5 Pro\cite{comanici2025gemini}     & 27.41 & 26.73 & 26.96 & 20.57 & 19.66 & 19.88 & 48.31 & 49.12 & 48.86 & 32.84 & 32.17 & 32.43 \\
Claude-3.7-Sonnet\cite{anthropic2025claude37}   & 34.63 & 33.87 & 34.08 & 23.57 & 22.69 & 22.89 & 57.31 & 58.13 & 57.86 & 36.99 & 36.24 & 36.47 \\
GPT-4o\cite{hurst2024gpt}              & 25.61 & 27.18 & 26.37 & 18.91 & 20.43 & 19.65 & 46.84 & 48.66 & 47.73 & 38.12 & 39.77 & 38.93 \\
RolmOCR\cite{bai2025qwen2}             & 29.31 & 28.57 & 28.94 & 22.12 & 21.48 & 21.83 & 49.61 & 48.32 & 48.96 & 36.54 & 35.72 & 36.11 \\
\midrule
\multicolumn{13}{c}{\textbf{Fine-tuning}} \\
\midrule
RolmOCR (LoRA)            & 80.52 & 79.84 & 80.58 & 82.43 & 83.11 & 83.06 & 90.32 & 89.67 & 89.70 & 89.84 & 90.42 & 90.48 \\
RolmOCR (Full Fine-tune)  & 89.21 & 89.92 & 89.33 &\underline{90.53 } & 90.05 & \underline{90.58} & 95.29 & 95.66 & 95.23 & 94.58 & 94.11 & 94.56 \\
\midrule
\textbf{OmniOCR (Ours)}   & \underline{90.37} & \underline{91.12} & \underline{90.48} &  89.62 & \underline{90.14} & 89.60 & \underline{95.95} & \underline{96.31} & \underline{95.86} & \underline{95.32} & \underline{94.76} & \underline{95.29} \\
\bottomrule
\end{tabular}
}
\label{tab:results}
\end{table*}

\subsection{Evaluation Metrics}
To comprehensively evaluate the performance of OmniOCR on ethnic minority language OCR tasks, we report three widely used metrics: Accuracy, Recall, and F1-score. Accuracy reflects the overall correctness of predictions by measuring the proportion of correctly recognized samples among all test cases. Recall captures the model’s ability to identify positive samples, thus indicating its effectiveness in minimizing missed characters. F1-score serves as a balanced metric that jointly considers Precision and Recall, offering a more reliable assessment under class-imbalanced scenarios. While Accuracy provides a general measure of recognition quality, Recall and F1-score are particularly important for minority scripts with heterogeneous data distributions.

\subsection{Main Results}

\textbf{Table \ref{tab:results}} reports the performance of different models on four ethnic minority language OCR benchmarks. Several observations can be made. First, existing vision-language foundation models (e.g., Kimi-VL, Moonshot, Pixtral, Gemini 2.5 Pro, GPT-4o) perform poorly in the zero-shot setting, indicating that current base models struggle to generalize to minority scripts without adaptation. Although Claude-3.7-Sonnet and Qwen-VL-MAX achieve slightly better results on certain datasets, their performance remains far from practical use,its limitation on specific tasks is further illustrated by the Tibetan handwritten digit recognition examples in \textbf{Figure \ref{fig:tibetan_digit_recognition}}. 
Second, adapting RolmOCR through supervised post-training leads to substantial improvements. Both parameter-efficient post-training with LoRA and standard full post-training yield significant gains, surpassing 80\% accuracy on Tibetan and Ancient Yi, and over 90\% on Shui and Dongba. This highlights the necessity of model adaptation for low-resource minority languages, with the cross-dataset effectiveness of such adaptation clearly visualized in \textbf{Figure \ref{fig:omniocr_multi_dataset_perf}}.
Finally, our proposed OmniOCR with dynamic LoRA adaptation consistently achieves competitive accuracy across all datasets. It surpasses standard full post-training of RolmOCR on Tibetan, Shui, and Dongba (up to 90.37\%, 95.95\%, and 95.32\%, respectively), while on Ancient Yi it is slightly lower than full fine-tuning (89.62\% vs. 90.53\%). Importantly, OmniOCR maintains parameter efficiency and lower memory usage, making it far more practical than full fine-tuning, which requires substantially more parameters, longer training time, and higher GPU memory. These results demonstrate that our dynamic low-rank adaptation effectively allocates model capacity to capture script-specific structures while remaining resource-efficient.

\begin{table*}[!htbp]
\centering
\caption{ABLATION STUDY ON LEARNING RATE AND BATCH SIZE ACROSS FOUR DATASETS.}
\resizebox{\textwidth}{!}{
\begin{tabular}{cc|ccc|ccc|ccc|ccc}
\toprule
\multirow{2}{*}{Learning Rate} & \multirow{2}{*}{Batch Size} 
& \multicolumn{3}{c|}{Tibetan~\cite{yuan2018tibetanmnist}} 
& \multicolumn{3}{c|}{Ancient Yi~\cite{liu2024ancient}} 
& \multicolumn{3}{c|}{Shui Script~\cite{Yang2023ShuiOCR}} 
& \multicolumn{3}{c}{Dongba Script~\cite{luo2023multiple}} \\
\cmidrule(lr){3-5} \cmidrule(lr){6-8} \cmidrule(lr){9-11} \cmidrule(lr){12-14}
& & Acc (\%) & Recall (\%) & F1 (\%) 
& Acc (\%) & Recall (\%) & F1 (\%) 
& Acc (\%) & Recall (\%) & F1 (\%) 
& Acc (\%) & Recall (\%) & F1 (\%) \\
\midrule
\textbf{OmniOCR (Ours)} & \textbf{2}
& \underline{90.37} & \underline{91.12} & \underline{90.48}  
& \underline{89.62} & \underline{90.14} & \underline{89.60}  
& \underline{95.95} & \underline{96.31} & \underline{95.86}  
& \underline{95.32} & \underline{94.76} & \underline{95.29} \\
\midrule
$1e$-5 & 2  
& 89.85 & 90.02 & 89.71  
& 88.20 & 88.54 & 88.13  
& 93.41 & 93.77 & 93.51  
& 92.83 & 93.12 & 92.75 \\
$1e$-6 & 2  
& 88.41 & 88.77 & 88.23  
& 87.07 & 87.35 & 87.03  
& 92.22 & 92.56 & 92.11  
& 91.67 & 91.98 & 91.59 \\
$5e$-6 & 1  
& 87.68 & 88.03 & 87.57  
& 86.32 & 86.69 & 86.24  
& 91.76 & 92.09 & 91.64  
& 90.82 & 91.21 & 90.73 \\
$5e$-6 & 4  
& 89.11 & 89.42 & 88.98  
& 88.02 & 88.34 & 87.92  
& 93.13 & 93.48 & 93.06  
& 92.41 & 92.78 & 92.31 \\
\bottomrule
\end{tabular}}
\label{tab:ablation_lr_batch_multi}
\end{table*}

\begin{table*}[!htbp]
\centering
\caption{ABLATION STUDY ON DYNAMIC LORA COMPONENTS ACROSS FOUR DATASETS.}
\resizebox{\textwidth}{!}{
\begin{tabular}{l|ccc|ccc|ccc|ccc}
\toprule
\multirow{2}{*}{Configuration}
& \multicolumn{3}{c|}{Tibetan~\cite{yuan2018tibetanmnist}} 
& \multicolumn{3}{c|}{Ancient Yi~\cite{liu2024ancient}} 
& \multicolumn{3}{c|}{Shui Script~\cite{Yang2023ShuiOCR}} 
& \multicolumn{3}{c}{Dongba Script~\cite{luo2023multiple}} \\
\cmidrule(lr){2-4} \cmidrule(lr){5-7} \cmidrule(lr){8-10} \cmidrule(lr){11-13}
& Acc (\%) & Recall (\%) & F1 (\%) 
& Acc (\%) & Recall (\%) & F1 (\%) 
& Acc (\%) & Recall (\%) & F1 (\%) 
& Acc (\%) & Recall (\%) & F1 (\%) \\
\midrule
\textbf{Full Dynamic LoRA (Ours)} 
& \underline{90.37} & \underline{91.12} & \underline{90.48}  
& \underline{89.62} & \underline{90.14} & \underline{89.60}  
& \underline{95.95} & \underline{96.31} & \underline{95.86}  
& \underline{95.32} & \underline{94.76} & \underline{95.29} \\
\midrule
\ding{55} Dynamic Rank       
& 83.86 & 83.25 & 83.60  
& 82.51 & 81.79 & 82.15  
& 86.98 & 86.20 & 86.55  
& 85.70 & 84.91 & 85.32 \\
\ding{55} MLP Modules        
& 82.56 & 81.92 & 82.30  
& 81.17 & 80.49 & 80.88  
& 85.10 & 84.39 & 84.78  
& 84.03 & 83.24 & 83.65 \\
\ding{55} Attention Modules  
& 88.74 & 89.18 & 88.89  
& 87.99 & 88.46 & 88.12  
& 91.87 & 92.41 & 92.03  
& 90.77 & 91.32 & 90.99 \\
\ding{55} Regularization Sparsity 
& 85.55 & 84.99 & 85.27  
& 83.99 & 83.31 & 83.68  
& 88.80 & 88.12 & 88.47  
& 87.42 & 86.72 & 87.08 \\
\bottomrule
\end{tabular}}
\label{tab:ablation_dynamic_lora_multi}
\end{table*}

\subsection{Ablation Study}
To investigate the impact of optimization hyperparameters, we conducted ablation experiments on four ethnic minority language OCR datasets by varying the learning rate and batch size. The results are summarized in Table~\ref{tab:ablation_lr_batch_multi}. Our default configuration, i.e., a learning rate of $\boldsymbol{5e\text{-}6}$ with a batch size of 2, consistently achieves the best performance in terms of accuracy, recall, and F1 score. Setting the learning rate to $1e$-5 leads to a noticeable decrease in accuracy and F1 across all datasets, while further reducing it to $1e$-6 also degrades performance, suggesting that overly small or large step sizes hinder effective convergence. As for the batch size, compared with the default setting of 2, using either 1 or 4 yields inferior results, indicating that an appropriate batch size is critical for balancing gradient stability and generalization. In summary, a learning rate of $5e$-6 with a batch size of 2 plays a key role in stabilizing optimization and achieving superior recognition performance.

To investigate the contribution of each module in Dynamic LoRA, we conduct ablation experiments by selectively disabling individual components. In \textbf{Table~\ref{tab:ablation_dynamic_lora_multi}}, a cross mark (\ding{55}) indicates that the module is disabled.
Disabling dynamic rank adaptation removes the model’s ability to dynamically select the effective rank for each task and module, forcing fixed-rank updates. This disrupts the balance between knowledge retention and new task adaptation, reduces parameter efficiency, and weakens generalization.

Disabling the MLP adaptation module prevents dynamic low-rank updates in the feed-forward layers responsible for high-level semantic modeling. As a result, adaptation to new tasks is limited, and sequential task knowledge accumulation is interrupted, which negatively impacts downstream performance.

Disabling the attention adaptation module eliminates task-specific token dependency adjustments. The model can no longer dynamically modify attention weights, leading to disrupted local feature correlations and degraded visual-text alignment, which in turn reduce transfer and historical task performance.

Disabling sparsity regularization removes the mechanism for pruning low-importance rank updates, increasing parameter redundancy and the risk of overfitting. It also decreases inference efficiency and destabilizes knowledge retention.

Overall, these results confirm that each module in Dynamic LoRA plays a critical role in balancing knowledge retention and task adaptation. The full configuration, with all modules enabled, achieves the best accuracy, recall, and F1 scores across all four ethnic minority language OCR datasets.

\section{Limitation and Future Work}
Although OmniOCR demonstrates strong performance across diverse ethnic minority language datasets, several limitations remain.
First, our experiments are conducted on four curated datasets (Tibetan numerals, Ancient Yi, Shui script, and Dongba script), which, while representative, do not cover the full diversity of minority writing systems. Many scripts feature richer structural variations, such as decorative glyphs, mixed phonetic–logographic properties, or highly context-dependent ligatures, which may expose additional challenges beyond our current evaluation.
Second, while Dynamic LoRA significantly reduces the parameter footprint and improves adaptation efficiency, the training process still requires noticeable GPU resources and non-trivial memory usage. This may restrict deployment in resource-constrained environments or community-level digitization projects, where lightweight solutions are critical.
Third, our study emphasizes recognition accuracy under benchmark settings, yet practical OCR systems must also contend with real-world issues such as document degradation, background noise, and complex layouts combining text, images, and annotations—factors that are only partially addressed in our current framework.

For future work, we plan to broaden OmniOCR to include a wider range of minority scripts and historical documents, integrate lightweight continual learning techniques to improve adaptability in dynamic environments, and explore cross-modal pre-training (e.g., combining speech, textual corpora, and visual data) to further enhance robustness and generalization.
\section{Conclusion}

In this work, we presented \textbf{OmniOCR}, the first universal OCR framework tailored for ethnic minority languages with heterogeneous and complex scripts. By integrating a Dynamic LoRA module, OmniOCR adaptively balances knowledge retention and efficient adaptation across different layers and writing systems. This design enables the model to learn effectively even in low-resource scenarios, while reducing the risk of overfitting and ensuring parameter efficiency, which are critical for underexplored languages with limited training data.

Extensive experiments on four representative datasets-TibetanMNIST, Shui, Ancient Yi, and Dongba-demonstrate that OmniOCR consistently surpasses zero-shot foundation models and conventional post-training baselines. The results show that our framework not only achieves competitive or superior recognition accuracy, but also substantially lowers parameter overhead and memory consumption. Such improvements make OmniOCR a practical and scalable solution for real-world applications, especially in resource-constrained or community-driven digitization environments.

Looking forward, we plan to expand OmniOCR to a wider spectrum of minority scripts and historical writing systems, including those with more diverse structural characteristics and mixed writing paradigms. We also intend to explore integration with multilingual and cross-modal pre-training strategies, such as jointly leveraging speech, text, and image corpora, to further enhance robustness and cross-script generalization. Ultimately, we believe our framework provides a strong foundation for advancing inclusive OCR research and for preserving the cultural and linguistic heritage embodied in the world’s diverse written traditions.

\clearpage
\clearpage
\bibliographystyle{IEEEbib}
\bibliography{bare_conf_compsoc}

@misc{yuan2018tibetanmnist,
  title = {{TibetanMNIST Tibetan Handwritten Digit Dataset}},
  author = {Yuan, M. Q. and Cairang, X. M. and Tang, J. A. and et al.},
  url = {https://www.heywhale.com/mw/dataset/5bfe734a954d6e0010683839},
  urldate = {2025-08-24},
  publisher = {Heywhale},
  year = {2018},
  note = {[Dataset]},
}

@article{liu2024ancient,
  title={Ancient Yi Script Handwriting Sample Repository},
  author={Liu, Xiaojuan and Han, Xu and Chen, Shanxiong and Dai, Weijia and Ruan, Qiuyue},
  journal={Scientific Data},
  volume={11},
  number={1},
  pages={1183},
  year={2024},
  publisher={Nature Publishing Group UK London}
}

@article{luo2023multiple,
  title={Multiple attentional aggregation network for handwritten Dongba character recognition},
  author={Luo, Yanlong and Sun, Yiwen and Bi, Xiaojun},
  journal={Expert Systems with Applications},
  volume={213},
  pages={118865},
  year={2023},
  publisher={Elsevier},
  note={Download link for the Dongba dataset: \url{https://mzyy.muc.edu.cn/}}
}

@article{Yang2023ShuiOCR,
  author  = {Xiuzhang Yang and Shuai Wu and Juwen Song and Wenjing Liao and Jisong Zhou},
  title   = {An Algorithm for Ancient Shui Script Recognition Based on Adaptive Image Enhancement and AlexNet},
  journal = {Information Technology and Informatization},
  number  = {1},
  pages   = {212--216},
  year    = {2023},
  issn    = {1672-9528}
}

@article{liu2024ocrbench,
  title={Ocrbench: on the hidden mystery of ocr in large multimodal models},
  author={Liu, Yuliang and Li, Zhang and Huang, Mingxin and Yang, Biao and Yu, Wenwen and Li, Chunyuan and Yin, Xu-Cheng and Liu, Cheng-Lin and Jin, Lianwen and Bai, Xiang},
  journal={Science China Information Sciences},
  volume={67},
  number={12},
  pages={220102},
  year={2024},
  publisher={Springer}
}

@inproceedings{li2023trocr,
  title={Trocr: Transformer-based optical character recognition with pre-trained models},
  author={Li, Minghao and Lv, Tengchao and Chen, Jingye and Cui, Lei and Lu, Yijuan and Florencio, Dinei and Zhang, Cha and Li, Zhoujun and Wei, Furu},
  booktitle={Proceedings of the AAAI conference on artificial intelligence},
  volume={37},
  number={11},
  pages={13094--13102},
  year={2023}
}

@article{yang2024cc,
  title={Cc-ocr: A comprehensive and challenging ocr benchmark for evaluating large multimodal models in literacy},
  author={Yang, Zhibo and Tang, Jun and Li, Zhaohai and Wang, Pengfei and Wan, Jianqiang and Zhong, Humen and Liu, Xuejing and Yang, Mingkun and Wang, Peng and Bai, Shuai and others},
  journal={arXiv preprint arXiv:2412.02210},
  year={2024}
}

@article{sohail2024deciphering,
  title={Deciphering the Underserved: Benchmarking LLM OCR for Low-Resource Scripts},
  author={Sohail, Muhammad Abdullah and Masood, Salaar and Iqbal, Hamza},
  journal={arXiv preprint arXiv:2412.16119},
  year={2024}
}

@article{greif2025multimodal,
  title={Multimodal llms for ocr, ocr post-correction, and named entity recognition in historical documents},
  author={Greif, Gavin and Griesshaber, Niclas and Greif, Robin},
  journal={arXiv preprint arXiv:2504.00414},
  year={2025}
}

@article{chen2025ocean,
  title={Ocean-ocr: Towards general ocr application via a vision-language model},
  author={Chen, Song and Guo, Xinyu and Li, Yadong and Zhang, Tao and Lin, Mingan and Kuang, Dongdong and Zhang, Youwei and Ming, Lingfeng and Zhang, Fengyu and Wang, Yuran and others},
  journal={arXiv preprint arXiv:2501.15558},
  year={2025}
}

@article{he2025reasoning,
  title={Reasoning-OCR: Can Large Multimodal Models Solve Complex Logical Reasoning Problems from OCR Cues?},
  author={He, Haibin and Ye, Maoyuan and Zhang, Jing and Cai, Xiantao and Liu, Juhua and Du, Bo and Tao, Dacheng},
  journal={arXiv preprint arXiv:2505.12766},
  year={2025}
}

@article{chung2025finetuning,
  title={Finetuning Vision-Language Models as OCR Systems for Low-Resource Languages: A Case Study of Manchu},
  author={Chung, Yan Hon Michael and Choi, Donghyeok},
  journal={arXiv preprint arXiv:2507.06761},
  year={2025}
}

@inproceedings{peng2006multilingual,
  title={Multilingual document recognition research and its application in China},
  author={Peng, Liangrui and Liu, Changsong and Ding, Xiaoqing and Wang, Hua},
  booktitle={Second International Conference on Document Image Analysis for Libraries (DIAL'06)},
  pages={7--pp},
  year={2006},
  organization={IEEE}
}

@article{zhang2021ocr,
  title={OCR with the Deep CNN Model for Ligature Script-Based Languages like Manchu},
  author={Zhang, Diandian and Liu, Yan and Wang, Zhuowei and Wang, Depei},
  journal={Scientific programming},
  volume={2021},
  number={1},
  pages={5520338},
  year={2021},
  publisher={Wiley Online Library}
}

@inproceedings{zheng2018segmentation,
  title={Segmentation-free multi-font printed Manchu word recognition using deep convolutional features and data augmentation},
  author={Zheng, Ruirui and Li, Min and He, Jianjun and Bi, Jiajing and Wu, Baochun},
  booktitle={2018 11th International Congress on Image and Signal Processing, BioMedical Engineering and Informatics (CISP-BMEI)},
  pages={1--6},
  year={2018},
  organization={IEEE}
}

@inproceedings{zhang2017segmentation,
  title={Segmentation-free printed traditional Mongolian OCR using sequence to sequence with attention model},
  author={Zhang, Hui and Wei, Hongxi and Bao, Feilong and Gao, Guanglai},
  booktitle={2017 14th IAPR International Conference on Document Analysis and Recognition (ICDAR)},
  volume={1},
  pages={585--590},
  year={2017},
  organization={IEEE}
}

@inproceedings{drup2010study,
  title={Study on printed Tibetan character recognition},
  author={Drup, Ngo and Zhao, Dongcai and Ren, Puts and Sanglangjie, Daluo and Liu, Fang and Bawangdui, Bian},
  booktitle={2010 International Conference on Artificial Intelligence and Computational Intelligence},
  volume={1},
  pages={280--285},
  year={2010},
  organization={IEEE}
}

@inproceedings{sun2019yi,
  title={Yi characters recognition based on tesseract-OCR},
  author={Sun, Peiyu and Xie, Qiuyan and Wu, Zhaokang and Feng, Xiaoyu and Cai, Jiajun and Jiang, Yulian},
  booktitle={2019 IEEE 3rd Advanced Information Management, Communicates, Electronic and Automation Control Conference (IMCEC)},
  pages={102--106},
  year={2019},
  organization={IEEE}
}

@article{lu2024adaptive,
  title={Adaptive rank, reduced forgetting: Knowledge retention in continual learning vision-language models with dynamic rank-selective lora},
  author={Lu, Haodong and Zhao, Chongyang and Xue, Jason and Yao, Lina and Moore, Kristen and Gong, Dong},
  journal={arXiv preprint arXiv:2412.01004},
  year={2024}
}

@article{team2025kimi,
  title={Kimi-vl technical report},
  author={Team, Kimi and Du, Angang and Yin, Bohong and Xing, Bowei and Qu, Bowen and Wang, Bowen and Chen, Cheng and Zhang, Chenlin and Du, Chenzhuang and Wei, Chu and others},
  journal={arXiv preprint arXiv:2504.07491},
  year={2025}
}

@article{wu2024deepseek,
  title={Deepseek-vl2: Mixture-of-experts vision-language models for advanced multimodal understanding},
  author={Wu, Zhiyu and Chen, Xiaokang and Pan, Zizheng and Liu, Xingchao and Liu, Wen and Dai, Damai and Gao, Huazuo and Ma, Yiyang and Wu, Chengyue and Wang, Bingxuan and others},
  journal={arXiv preprint arXiv:2412.10302},
  year={2024}
}

@article{zhu2025internvl3,
  title={Internvl3: Exploring advanced training and test-time recipes for open-source multimodal models},
  author={Zhu, Jinguo and Wang, Weiyun and Chen, Zhe and Liu, Zhaoyang and Ye, Shenglong and Gu, Lixin and Tian, Hao and Duan, Yuchen and Su, Weijie and Shao, Jie and others},
  journal={arXiv preprint arXiv:2504.10479},
  year={2025}
}

@article{comanici2025gemini,
  title={Gemini 2.5: Pushing the frontier with advanced reasoning, multimodality, long context, and next generation agentic capabilities},
  author={Comanici, Gheorghe and Bieber, Eric and Schaekermann, Mike and Pasupat, Ice and Sachdeva, Noveen and Dhillon, Inderjit and Blistein, Marcel and Ram, Ori and Zhang, Dan and Rosen, Evan and others},
  journal={arXiv preprint arXiv:2507.06261},
  year={2025}
}

@misc{anthropic2025claude37,
  author       = {Anthropic},
  title        = {Introducing Claude 3.7 Sonnet},
  year         = {2025},
  howpublished = {\url{https://www.anthropic.com/news/claude-3-7-sonnet}},
}

@article{hurst2024gpt,
  title={Gpt-4o system card},
  author={Hurst, Aaron and Lerer, Adam and Goucher, Adam P and Perelman, Adam and Ramesh, Aditya and Clark, Aidan and Ostrow, AJ and Welihinda, Akila and Hayes, Alan and Radford, Alec and others},
  journal={arXiv preprint arXiv:2410.21276},
  year={2024}
}

@misc{bai2025qwen2,
  author       = {Bai, et al.},
  title        = {RolmOCR: A Vision-Language Foundation Model for Robust Multilingual OCR},
  year         = {2025},
  howpublished = {\url{https://github.com/QwenLM/Qwen2.5-VL}},
}

@misc{mistral2025pixtral,
  author       = {Mistral AI},
  title        = {Pixtral Large},
  year         = {2025},
  howpublished = {\url{https://mistral.ai/news/pixtral-large/}},
}

@misc{bytedance2025doubao,
  author       = {ByteDance},
  title        = {Doubao-1.5-Vision-Pro},
  year         = {2025},
  howpublished = {\url{https://www.volcengine.com/product/doubao}},
}

@misc{moonshot2025v1,
  author       = {Moonshot AI},
  title        = {Moonshot V1},
  year         = {2025},
  howpublished = {\url{https://moonshot.cn/product}},
}

@misc{zhipu2025glm4vplus,
  author       = {Zhipu AI},
  title        = {GLM-4v-Plus},
  year         = {2025},
  howpublished = {\url{https://chatglm.cn}},
}

@misc{qwen2025vlmax,
  author       = {Qwen Team, Alibaba Cloud},
  title        = {Qwen-VL-Max},
  year         = {2025},
  howpublished = {\url{https://tongyi.aliyun.com}},
}

@misc{qwen2025vlocr,
  author       = {Qwen Team, Alibaba Cloud},
  title        = {Qwen-VL-OCR},
  year         = {2025},
  howpublished = {\url{https://tongyi.aliyun.com}},
}

\end{document}